\documentclass[letterpaper, 10 pt, journal, twoside]{IEEEtran}
\usepackage{graphics} 
\graphicspath{{figures/}}\usepackage{epsfig} 
\usepackage{mathptmx} 
\usepackage{amsmath} 
\usepackage{amssymb}  

\usepackage{cite} 

\usepackage[colorlinks,linkcolor=red]{hyperref} 
\usepackage{color}
\usepackage{multirow}
\usepackage{soul}

\usepackage{balance} 

\begin{document}

\title{RGB-D Inertial Odometry for a Resource-restricted Robot in Dynamic Environments}

\author{Jianheng Liu$^{1}$, Xuanfu Li$^{2}$, Yueqian Liu$^{1}$, and Haoyao Chen$^{1}$%
\thanks{Manuscript received: February 25, 2022; Revised: May 26, 2022; Accepted: June 28, 2022 .}
\thanks{This paper was recommended for publication by Editor Javier Civera upon evaluation of the Associate Editor and Reviewers' comments.
This work was supported in part by the National Natural Science	Foundation of China (Grant No.U21A20119 and No.U1713206) and in part by the Shenzhen Science and Innovation Committee (Grant No.JCYJ20200109113412326; No.JCYJ20210324120400003; No.JCYJ20180507183837726; No.JCYJ20180507183456108).}
\thanks{$^{1}$J.H. Liu, Y.Q. Liu and H.Y. Chen* are with the School of Mechanical Engineering and Automation, Harbin Institute of Technology Shenzhen, P.R. China (* represents the corresponding author).
        {\tt\footnotesize \{liujianheng, liuyueqian\}@stu.hit.edu.cn, hychen5@hit.edu.cn}}%
\thanks{$^{2}$X.F. Li is with the Department of HiSilicon Research, Huawei Technology Co., Ltd, P.R. China.
        {\tt\footnotesize lixuanfu@huawei.com}}%
\thanks{Digital Object Identifier (DOI): see top of this page.}
}
%
%

\markboth{IEEE Robotics and Automation Letters. Preprint Version. Accepted June, 2022}
{Liu \MakeLowercase{\textit{et al.}}: RGB-D Inertial Odometry for a Resource-restricted Robot in Dynamic Environments} 

%



\maketitle

\begin{abstract}
  Current simultaneous localization and mapping (SLAM) algorithms perform well in static environments but easily fail in dynamic environments.
  Recent works introduce deep learning-based semantic information to SLAM systems to reduce the influence of dynamic objects. 
  However, it is still challenging to apply a robust localization in dynamic environments for resource-restricted robots.
  This paper proposes a real-time RGB-D inertial odometry system for resource-restricted robots in dynamic environments named Dynamic-VINS.
  Three main threads run in parallel: object detection, feature tracking, and state optimization. 
  The proposed Dynamic-VINS combines object detection and depth information for dynamic feature recognition and achieves performance comparable to semantic segmentation. 
  Dynamic-VINS adopts grid-based feature detection and proposes a fast and efficient method to extract high-quality FAST feature points.
  IMU is applied to predict motion for feature tracking and moving consistency check.
  The proposed method is evaluated on both public datasets and real-world applications and shows competitive localization accuracy and robustness in dynamic environments.
  Yet, to the best of our knowledge, it is the best-performance real-time RGB-D inertial odometry for resource-restricted platforms in dynamic environments for now. 
  The proposed system is open source at: \url{https://github.com/HITSZ-NRSL/Dynamic-VINS.git}
\end{abstract}


%
\IEEEpeerreviewmaketitle

\section{Introduction}
%
%
%
%
\IEEEPARstart{S}{imultaneous} localization and mapping (SLAM) is a foundational capability for many emerging applications, such as autonomous mobile robots and augmented reality.
Cameras as portable sensors are commonly equipped on mobile robots and devices.
Therefore, visual SLAM (vSLAM) has received tremendous attention over the past decades.
Lots of works\cite{engelDirectSparseOdometry2018, vinsmono, genevaOpenVINSResearchPlatform2020,mur-artalORBSLAM2OpenSourceSLAM2017} are proposed to improve visual SLAM systems' performance.
Most of the existing vSLAM systems depend on a static world assumption.
Stable features in the environment are used to form a solid constraint for Bundle Adjustment\cite{bundleadjustment}.
However, in real-world scenarios like shopping malls and subways, dynamic objects such as moving people, vehicles, and unknown objects, have an adverse impact on pose optimization.
Although some approaches like RANSAC\cite{fischler1981ransac} can suppress the influence of dynamic features to a certain extent, it will become overwhelmed when a vast number of dynamic objects appear in the scene.

Therefore, it is necessary for the system to reduce dynamic objects' influence on the estimation results consciously.
The pure geometric methods\cite{sunImprovingRGBDSLAM2017,palazzoloReFusion3DReconstruction2019,daiRGBDSLAMDynamic2022} are widely used to handle dynamic objects, but it is unable to cope with latent or slightly moving objects.
With the development of deep learning, many researchers have tried combining multi-view geometric methods with semantic information\cite{liuSSDSingleShot2016, redmonYOLOv3IncrementalImprovement2018,badrinarayananSegNetDeepConvolutional2017, heMaskRCNN2017} to implement a robust SLAM system in dynamic environments.
To avoid the accidental deletion of stable features through object detection\cite{xiaoDynamicSLAMSemanticMonocular2019}, recent dynamic SLAM systems\cite{yuDSSLAMSemanticVisual2018,bescosDynaSLAMTrackingMapping2018} exploit the advantages of pixel-wise semantic segmentation for a better recognition of dynamic features.
Due to the expensive computing resource consumption of semantic segmentation, it is difficult for a semantic-segmentation-based SLAM system to run in real-time.
Therefore, some researchers have tried to perform semantic segmentation only on keyframes and track moving objects via moving probability propagation\cite{zhongDetectSLAMMakingObject2018,liuRDSSLAMRealTimeDynamic2021} or direct method\cite{ballesterDOTDynamicObject2021a} on each frame.
In the cases of missed detections or object tracking failures, the pose optimization is imprecise.
Moreover, since semantic segmentation is performed after keyframe selection, real-time precise pose estimation is inaccessible, and unstable dynamic features in the original frame may also cause redundant keyframe creation and unnecessary computational burdens.
  
\begin{figure*}[t]
  \begin{center}
    \includegraphics[width = 1.0\textwidth]{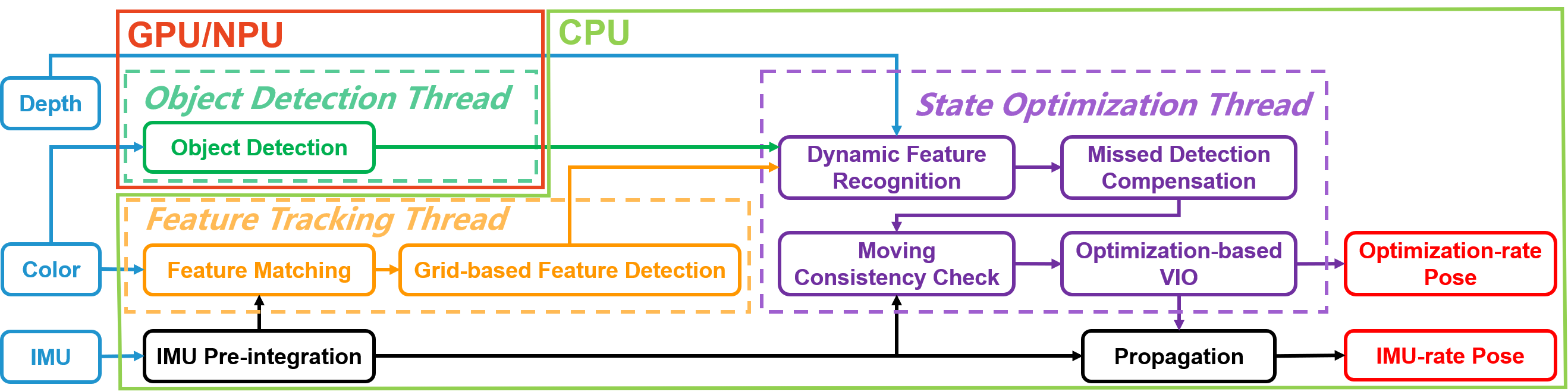}
  \end{center}
  \caption{{The framework of Dynamic-VINS. 
    The contributing modules are highlighted and surrounded by dash lines with different colors.
    Three main threads run in parallel in Dynamic-VINS.
    Features are tracked and detected in the feature tracking thread.
    The object detection thread detects dynamic objects in each frame in real-time.
    The state optimization thread summarizes the features information, object detection results, and depth image to recognize the dynamic features.
    Finally, stable features and IMU preintegration results are used for pose estimation.}}
  \label{fig:system_pipeline}
\end{figure*}

The above systems still require too many computing resources to perform robust real-time localization in dynamic environments for Size, Weight, and Power (SWaP) restricted mobile robots or devices.
Some researchers\cite{schauweckerMarkerlessVisualControl2012,zakaryaienejadARMVOEfficientMonocular2019,bahnamStereoVisualInertial2021} try to run visual odometry in real-time on embedded computing devices, yet the keyframe-based visual odometry is not performed\cite{younesKeyframebasedMonocularSLAM2017}, which makes their accuracy unsatisfactory.
At the same time, increasingly embedded computing platforms are equipped with NPU/GPU computing units, such as HUAWEI Atlas200, NVIDIA Jetson, etc.
It enables lightweight deep learning networks to run on the embedded computing platform in real-time.
Some studies\cite{xiaoDynamicSLAMSemanticMonocular2019,jiRealtimeSemanticRGBD2021} implemented a keyframe-based dynamic SLAM system running on embedded computing platforms.
However, these works are still difficult to balance efficiency and accuracy for mobile robot applications.

To address all these issues, this paper proposes a real-time RGB-D inertial odometry for resource-restricted robots in dynamic environments named Dynamic-VINS.
It enables edge computing devices to provide instant robust state feedback for mobile platforms with little computation burden. 
An efficient dynamic feature recognition module that does not require a high-precision depth camera can be used in mobile devices equipped with depth-measure modules.
The main contributions of this paper are as follows:
\begin{enumerate}
  \item 
  {An efficient optimization-based RGB-D inertial odometry is proposed to provide real-time state estimation results for resource-restricted robots in dynamic and complex environments.}
  \item 
  {Lightweight feature detection and tracking are proposed to cut the computing burden.
  In addition, dynamic feature recognition modules combining object detection and depth information are proposed to provide robust dynamic feature recognition in complex and outdoor environments.}
  \item 
  Validation experiments are performed to show the proposed system's competitive accuracy, robustness, and efficiency on resource-restricted platforms in dynamic environments.
\end{enumerate}

%
\section{SYSTEM OVERVIEW}\label{systemoverview}
	
The proposed SLAM system in this paper is extended based on VINS-Mono\cite{vinsmono} and VINS-RGBD\cite{vinsrgbd}; our framework is shown in Fig.~\ref{fig:system_pipeline}, and the contributing modules are highlighted with different colors. 
For efficiency, three main threads (surrounded by dash lines) run parallel in Dynamic-VINS: object detection, feature tracking, and state optimization.
Color images are passed to both the object detection thread and the feature tracking thread.
{IMU measurements between two consecutive frames are preintegrated\cite{imupreintegration} for feature tracking, moving consistency check, and state optimization.}

In the feature tracking thread, features are tracked with the help of IMU preintegration and detected by grid-based feature detection.
The object detection thread detects dynamic objects in each frame in real-time.
Then, the state optimization thread will summarize the features information, object detection results, and depth image to recognize the dynamic features.
A missed detection compensation module is conducted in case of missed detection.
The moving consistency check procedure combines the IMU preintegration and historical pose estimation results to identify potential dynamic features.
Finally, stable features and IMU preintegration results are used for the pose estimation.
And the propagation of the IMU is responsible for an IMU-rate pose estimation result.
Loop closure is also supported in this system, but this paper pays more attention to the localization independent of loop closure. 

\section{METHODOLOGY}\label{methodology}

This study proposes lightweight, high-quality feature tracking and detection methods to accelerate the system.
Semantic and geometry information from the input RGB-D images and IMU preintegration are applied for dynamic feature recognition and moving consistency check.
The missed detection compensation module plays a subsidiary role to object detection in case of missed detection.
Dynamic features on unknown objects are further identified by moving consistency check.
The proposed methods are divided into five parts for a detailed description.

\subsection{Feature Matching}\label{sec:feature_tracking}

\begin{figure}[htpb]
  \centering
  \includegraphics[width = 0.48\textwidth]{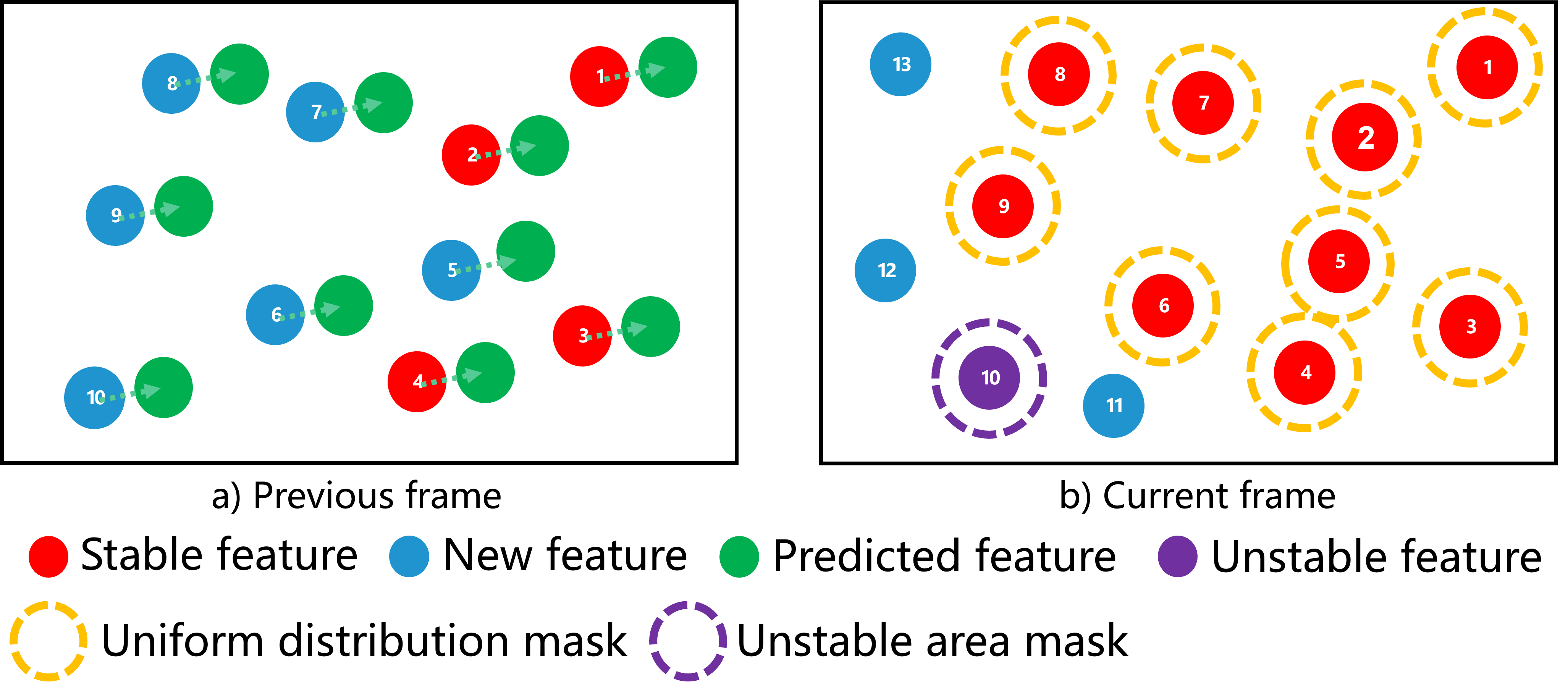}
  \caption{
    Illustration of feature tracking and detection.
    Stable features and new features are colored red and blue, respectively.
    The green circles denote the prediction for optical flow.
    The successfully tracked features turn red; otherwise, the features turn purple.
    The orange and purple dash-line circles as masks are set for a uniform feature distribution and reliable feature detection.
    New feature points are detected from unmasked areas in the current frame.
  }
  \label{fig:unstable_feature}
\end{figure}

For each incoming image, the feature points are tracked using the KLT sparse optical flow method \cite{lucas1981iterative}.
In this paper, the IMU measurements between frames are used to predict the motion of features.
Better initial position estimation of features is provided to improve the efficiency of feature tracking by reducing optical flow pyramid layers.
It can effectively discard unstable features such as noise and dynamic features with inconsistent motion.
The basic idea is illustrated in Fig.~\ref{fig:unstable_feature}.

In the previous frame, stable features are colored red, and newly detected features are colored blue.
When the current frame arrives, the IMU measurements between the current and previous frames are used to predict the feature position (green) in the current frame.
Optical flow uses the predicted feature position as the initial position to look for a match feature in the current frame.
The successfully tracked features are turned red, while those that failed to be tracked are marked as unstable features (purple).
In order to avoid the repetition and aggregation of feature detection, an orange circular mask centered on the stable feature is set; the region where the unstable features are located is considered an unstable feature detection region and masked with a purple circular to avoid unstable feature detection.
According to the mask, new features are detected from unmasked areas in the current frame and colored blue.

The above means can obtain uniformly distributed features to capture comprehensive constraints and avoid repeatedly extracting unstable features on the area with blurs or weak textures.
Long-term feature tracking can reduce the time consumption with the help of grid-based feature detection in the following.

\subsection{Grid-based Feature Detection}\label{sec:feature_detect}

The system maintains a minimum number of features for stability.
Therefore, feature points need to be extracted from the frame constantly.
This study adopts grid-based feature detection.
Image is divided into grids, and the boundary of each grid is padded to prevent the features at the edge of the grid from being ignored; the padding enables the current grid to obtain adjacent pixel information for feature detection.
Unlike traversing the whole image to detect features, only {the grid with insufficient matched features} will conduct feature detection.
The grid cell that fails to detect features due to weak texture or is covered by the mask will be skipped in the next detection frame to avoid repeated useless detection.
The thread pool technique is used to exploit the parallel performance of grid-based feature detection.
Thus, the time consumption of feature detection is significantly reduced without loss.

The FAST feature detector\cite{fast} can efficiently extract feature points but easily treats noise as features and extracts similar clustered features.
{Therefore, the ideas of mask in Sec.~\ref{sec:feature_tracking} and Non-Maximum-Suppression are combined to select high-quality and uniformly distributed FAST features.}

\begin{figure}[h]
  \centering
  \includegraphics[width = 0.48\textwidth]{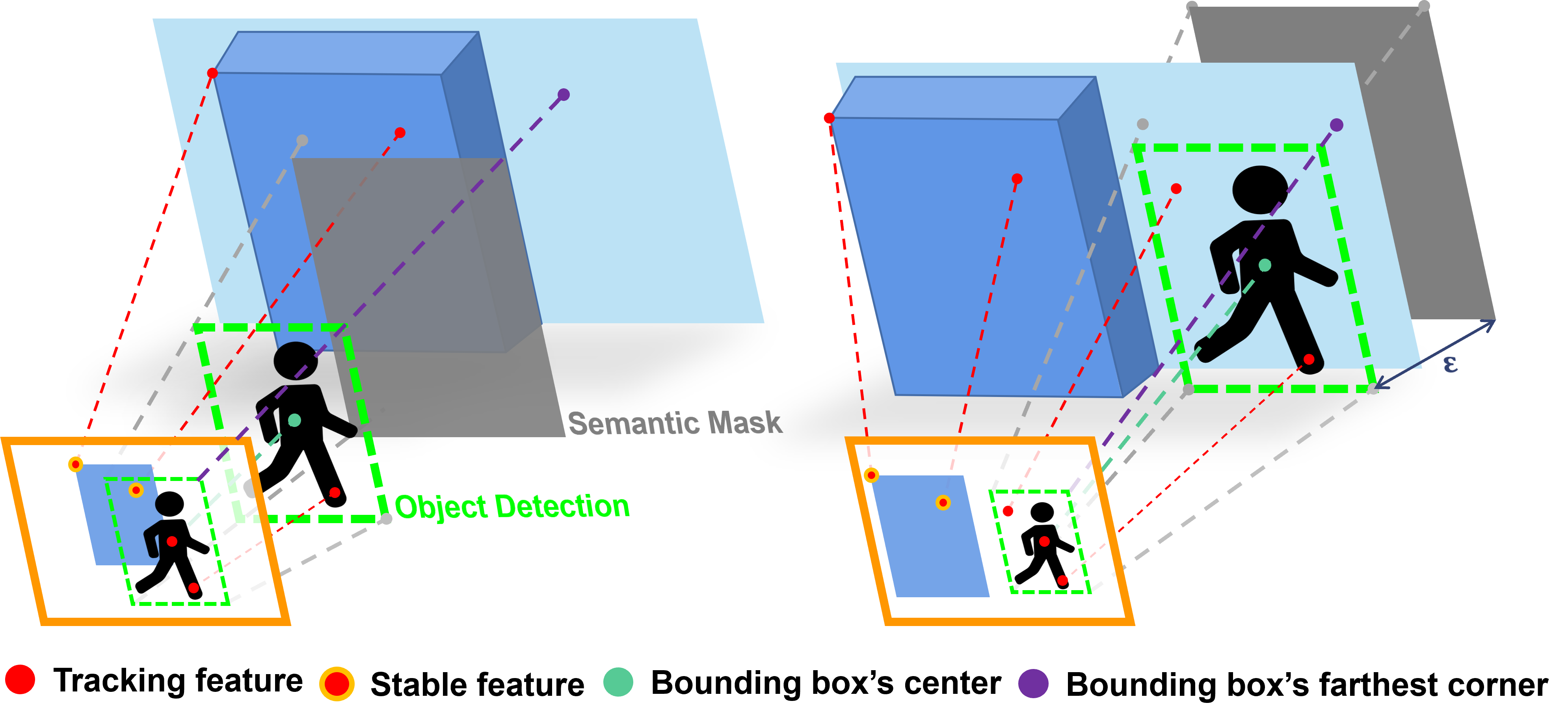}
  \caption{
    Illustration of semantic mask setting for dynamic feature recognition when all pixel's depth is available ($d>0$).
    The left scene represents when an objected bounding box's farthest corner's depth is bigger than the center to a threshold $\epsilon$ and a semantic mask with weighted depth is set between them to separate features on dynamic objects from the background. 
    Otherwise, the semantic mask is set behind the bounding box's center with the distance of $\epsilon$, shown on the right.
  }
  \label{fig:set_semantic_mask}
\end{figure}

\begin{figure*}[!t]
  \begin{center}
    \includegraphics[width = 1.0\textwidth]{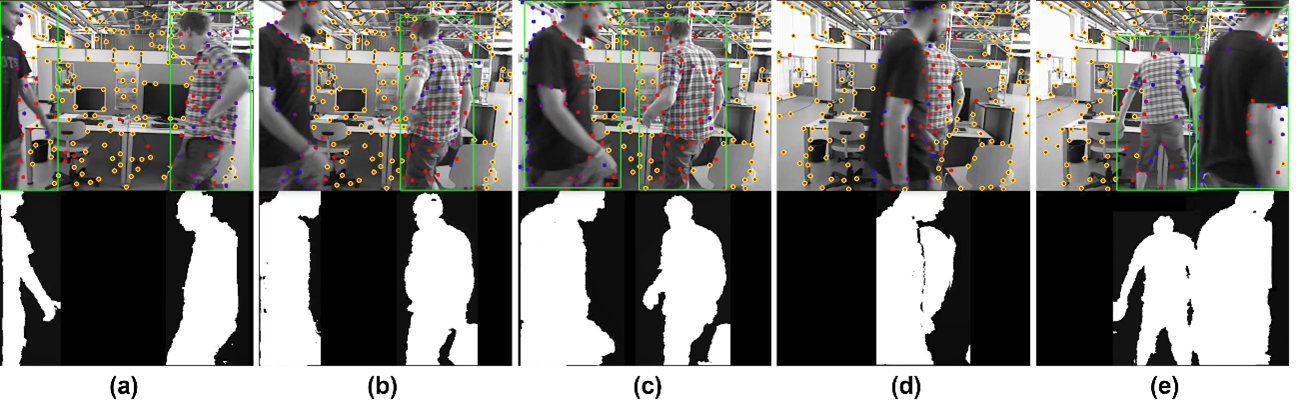}
  \end{center}
  \caption{
    Results of missed detection compensation.
    The dynamic feature recognition results are shown in the first row.
    The green box shows the dynamic object's position from the object detection results.
    The second row shows the generated semantic mask.
    With the help of missed detection compensation, even if object detection failed {in (b) and (d)}, a semantic mask including all dynamic objects could be built.
  }
  \label{fig:mdc}
\end{figure*}

\subsection{Dynamic Feature Recognition}

Most feature points can be stably tracked through the above improvement.
However, long-term tracking features on dynamic objects always come with abnormal motion and introduce wrong constraints to the system.
For the sake of efficiency and computational cost, a real-time single-stage object detection method, YOLOv3\cite{redmonYOLOv3IncrementalImprovement2018}, {is used to detect many kinds of dynamic scene elements like people and vehicles.
If a detected bounding box covers a large region of the image, blindly deleting feature points in the bounding box might result in no available features to provide constraints.
Therefore, semantic-segmentation-like masks are helpful to maintain the system's running by tracking features not occluded by dynamic objects.}

\begin{figure}[htbp]
  \centering
  \begin{minipage}{0.235\textwidth}
    \centering
    \includegraphics[width=\textwidth]{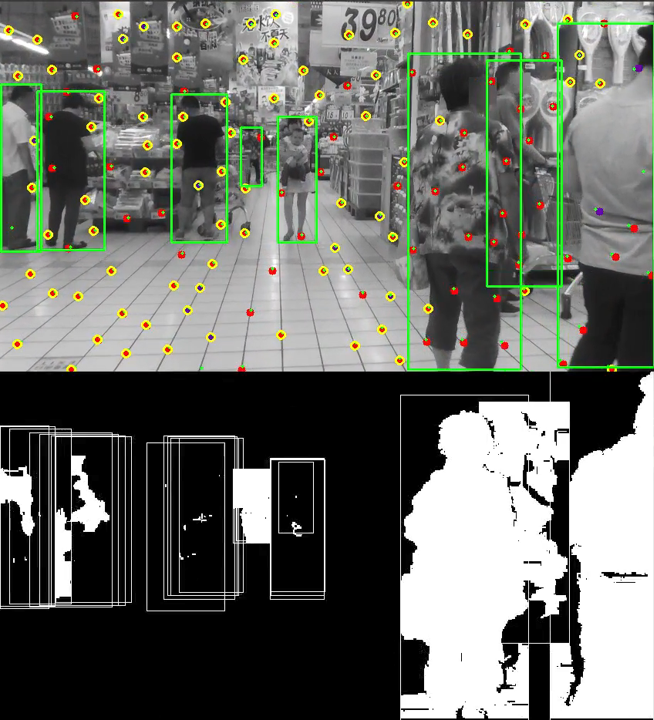}
    \footnotesize{(a) Seed Filling}
  \end{minipage}
  \begin{minipage}{0.235\textwidth}
    \centering
    \includegraphics[width=\textwidth]{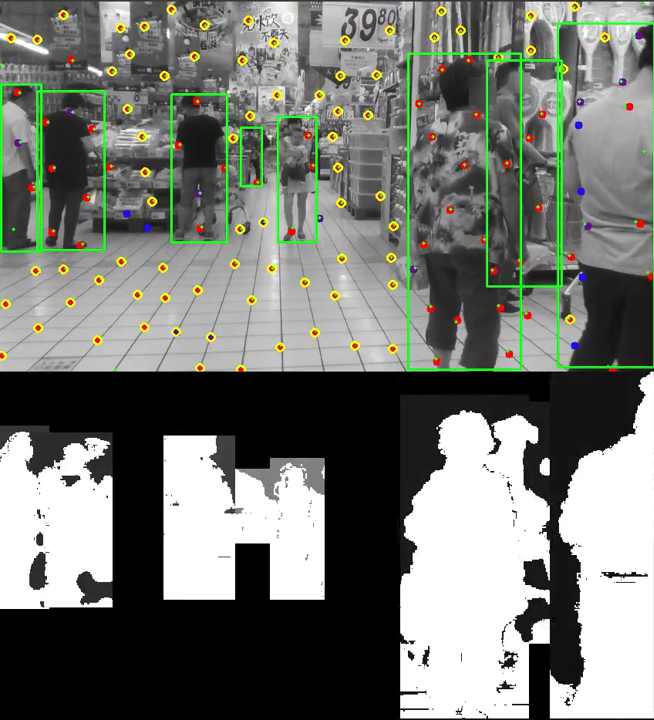}
    \footnotesize{(b) Proposed Method}
  \end{minipage}
  \caption{
    Results of dynamic feature recognition.
    The stable features are circled by yellow.
    The dynamic feature recognition results generated by Seed Filling and the proposed method are shown in (a) and (b), respectively. 
    The weighted depth $\bar{d}$ is colored gray; the brighter means a bigger value.
    The feature point on the white area will be marked as a dynamic feature.
  }
  \label{fig:dynamic_feature_recognition}
\end{figure}

This paper combines object detection and depth information for highly efficient dynamic feature recognition to achieve performance comparable to semantic segmentation.
As the farther the depth camera measures, the worse the accuracy is.
This problem makes some methods, such as Seed Filling, DBSCAN, and K-Means, which make full use of the depth information, exhibit poor performance with a low accuracy depth camera, as shown in Fig.~\ref{fig:dynamic_feature_recognition}(a).
Therefore, a set of points in the detected bounding box and depth information are integrated to obtain comparable performance to the semantic segmentation, as illustrated in Fig.~\ref{fig:set_semantic_mask}.

A pixel's depth $d$ is available, if $d>0$, otherwise, $d=0$.
{Considering that the bounding box corners of most dynamic objects correspond to the background points, and the dynamic objects commonly have a relatively large depth gap with the background. The $K$-th dynamic object's largest background depth ${}^{K}d_{max}$ is obtained as follow}
\begin{equation}
  {}^{K}d_{max} =\max\left({}^{K}d_{tl} , {}^{K}d_{tr} , {}^{K}d_{bl} , {}^{K}d_{br}\right),
\end{equation}
where ${}^{K}d_{tl}$, ${}^{K}d_{tr}$, ${}^{K}d_{bl}$, ${}^{K}d_{br}$ are the depth values of the $K$th object detection bounding box's corners, respectively.
Next, the $K$th {bounding box's depth threshold ${}^{K}\bar{d}$} is defined as
\begin{equation}
  {}^{K}\bar{d}=
  \begin{cases}
    \frac{1}{2}\left({}^{K}d_{\max } + {}^{K}d_{c}\right), & \text { if } {}^{K}d_{\max } - {}^{K}d_{c}>\epsilon,\ {}^{K}d_{c}>0,\\
    {}^{K}d_{c} + \epsilon, & \text { if } {}^{K}d_{\max } - {}^{K}d_{c}<\epsilon,\ {}^{K}d_{c}>0,\\
    {}^{K}d_{\max }, & \text { if } {}^{K}d_{\max }>0,\ {}^{K}d_{c}=0,\\
    +\infty, & \text { otherwise },
  \end{cases}
\end{equation}
where ${}^{K}d_{c}$ is the depth value of the bounding box's center;
$\epsilon>0$ is a predefined distance according to the most common dynamic objects' size in scenes. The depth threshold ${}^{K}\bar{d}$ is defined in the middle of the center's depth ${}^{K}d_{c}$ and the deepest background depth ${}^{K}d_{\max}$. When the dynamic object has a close connection with the background or is behind an object ${}^{K}d_{\max } - {}^{K}d_{c}<\epsilon$, the depth threshold is defined at $\epsilon$ distance from the dynamic object. If the depth is unavailable, a conservative strategy is adopted to choose an infinite depth as the threshold.

{On the semantic mask, the area covered by the $K$-th dynamic object bounding box is set to the weighted depth ${}^{K}\bar d$; the area without dynamic objects is set to 0.
Each incoming feature's depth $d$ is compared with the corresponding pixel's depth threshold $\bar{d}$ on the semantic mask. If $d<\bar{d}$, the feature is considered as a dynamic one.
Otherwise, the feature is considered as a stable one.
Therefore, the region where the depth value is smaller than the weighted depth $\bar d$ constitutes the generalized semantic mask, as shown in Fig.~\ref{fig:mdc} and Fig.~\ref{fig:dynamic_feature_recognition}(b).}

Considering that dynamic objects may exist in the field of view for a long time, the dynamic features are tracked but not used for pose estimation, different from directly deleting dynamic features.
According to its recorded information, each incoming feature point from the feature tracking thread will be judged whether it is a historical dynamic feature or not.
The above methods can avoid blindly deleting feature points while ensuring efficiency.
It can save time from detecting features on dynamic objects, has the robustness to the missed detection of object detection, and recycle false-positive dynamic features, as illustrated in Sec.~\ref{sec:mcc}.

\begin{figure*}[!t]
  \begin{center}
    \includegraphics[width = 1.0\textwidth]{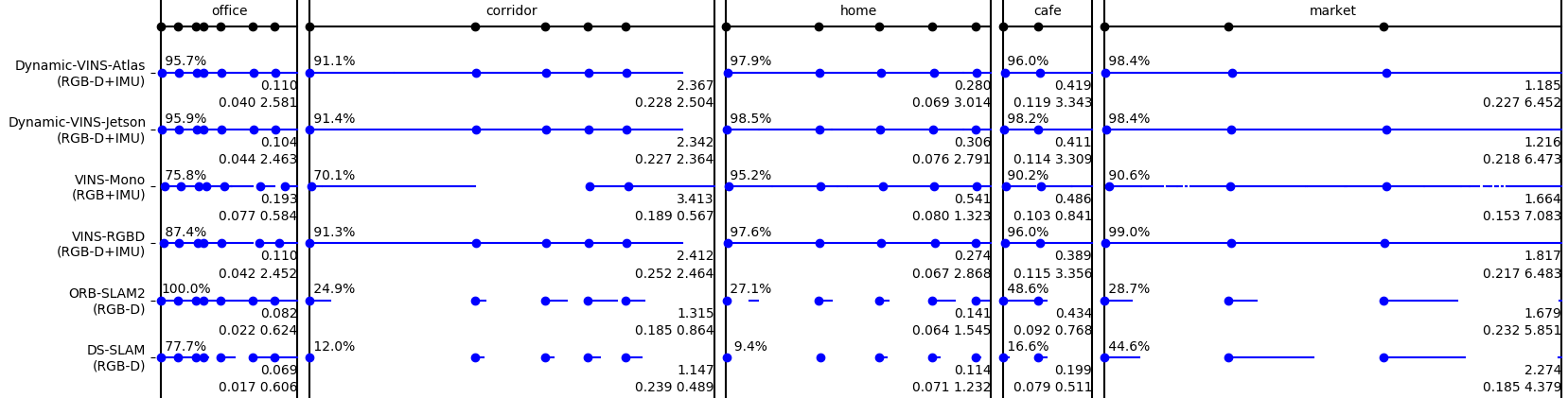}
  \end{center}
  \caption{Per-sequence testing results with the OpenLORIS-Scene datasets. Each black dot on the top line represents the start of one data sequence. For each algorithm, blue dots indicate successful initialization {moments}, and blue lines indicate successful tracking {span}. The percentage value on the top left of each scene is the average correct rate; the higher the correct rate of an algorithm, the more robust it is. The float value on the {first line below} is average ATE RMSE {and the values on the second line below are T.RPE and R.RPE from left to right}, and smaller means more accurate.
  }
  \label{fig:openloris}
\end{figure*}

\subsection{Missed Detection Compensation}

Since object detection might sometimes fail, the proposed Dynamic-VINS utilizes the previous detection results to predict the following detection result to compensate for missed detections.
It is assumed that the dynamic objects in adjacent frames have a consistent motion.
Once a dynamic object is detected, its pixel velocity and bounding box will be updated.
Assumed that $j$ is the current detected frame and $j-1$ is the previous detected frame, the pixel velocity {${}^{K}\mathbf{v}^{c_j}$ (pixel/frame)} of the $K$th dynamic object between frames is defined as 
\begin{equation}
  {{}^{K}\mathbf{v}^{c_j} =
  {}^{K}\mathbf{u}_c^{c_j}-{}^{K}\mathbf{u}_c^{c_{j-1}},}
\end{equation}
where ${}^{K}\mathbf{u}_c^{c_j}$, ${}^{K}\mathbf{u}_c^{c_{j-1}}$ represent the pixel location of the $K$th
object detection bounding box’s {center} in $j$th frame and $j-1$th frame, respectively. 
{A weighted predicted velocity ${}^{K}\mathbf{\hat{v}}$ is defined as}
\begin{equation}
  {{}^{K}\mathbf{\hat{v}}^{c_{j+1}} =\frac{1}{2}({}^{K}\mathbf{v}^{c_j}+{}^{K}\mathbf{\hat{v}}^{c_j}),}
\end{equation}
With the update going on, {the velocities of older frames will have a lower weight in ${}^{K}\mathbf{\hat{v}}$.}
If the object fail to be detected in the next frame, the bounding box ${}^{K}\mathbf{Box}$ containing the corners' pixel locations ${}^{K}\mathbf{u}_{tl},{}^{K}\mathbf{u}_{tr},{}^{K}\mathbf{u}_{bl}$ and ${}^{K}\mathbf{u}_{br}$, will be {updated based on the predicted velocity ${}^{K}\mathbf{\hat{v}}$} as follow
\begin{equation}
  {}^{K}\hat{\mathbf{Box}}^{c_{j+1}} =
  {}^{K}\mathbf{Box}^{c_{j}}+{}^{K}\mathbf{\hat{v}}^{c_{j+1}},
\end{equation}
When the missed detection time is over a threshold, this dynamic object's compensation will be abandoned.
The result is shown in Fig.~\ref{fig:mdc}.
It improves the recall rate of object detection and is helpful for a more consistent dynamic feature recognition.

\begin{figure}[h]
  \centering
  \begin{minipage}{0.2\textwidth}
    \centering
    \includegraphics[width=\textwidth]{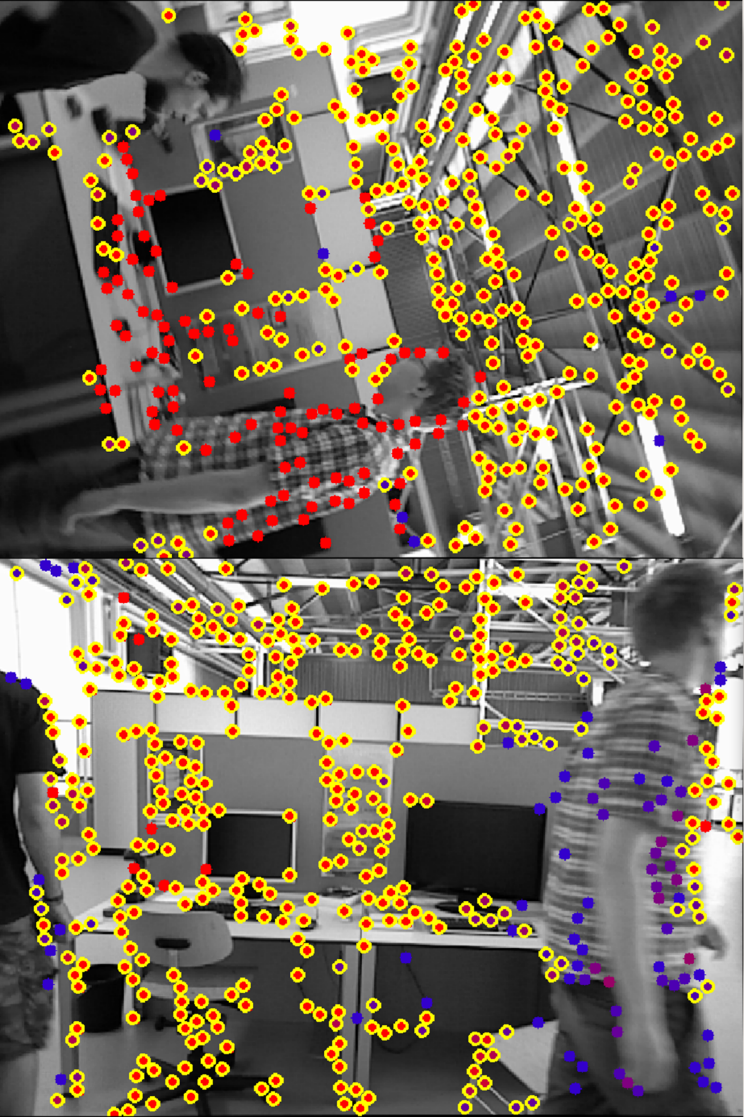}
  \end{minipage}
  \begin{minipage}{0.265\textwidth}
    \centering
    \includegraphics[width=\textwidth]{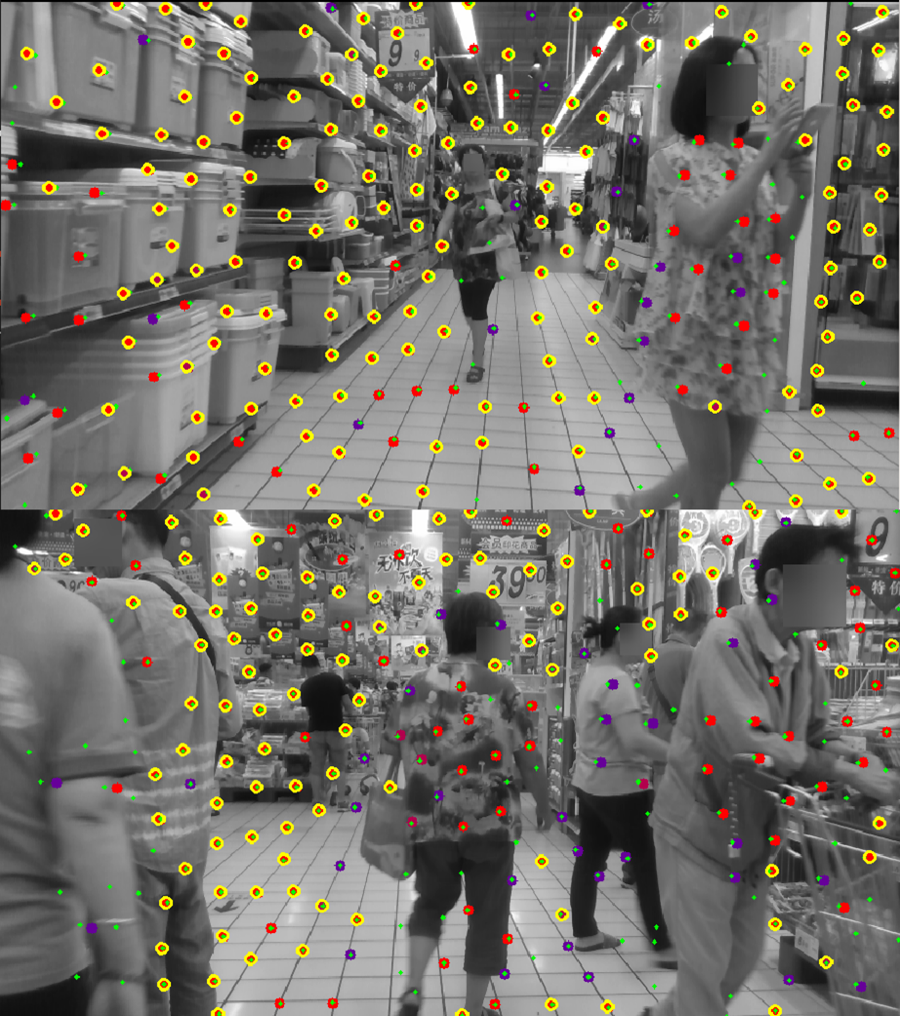}
  \end{minipage}
  \caption{
    {Results of Moving Consistency Check. Features without yellow circular are the outliers marked by the Moving Consistency Check module.}
  }
  \label{fig:mcc}
\end{figure}

\subsection{Moving Consistency Check}
\label{sec:mcc}

Since object detection can only recognize artificially defined dynamic objects and has a missed detection problem, the state optimization will still be affected by unknown moving objects like books moved by people.
Dynamic-VINS combines the pose predicted by IMU and the optimized pose in the sliding windows to recognize dynamic features.

Consider the $k$th feature is first observed in the $i$th image and is observed by other $m$ images in sliding windows.
The average reprojection residual $r_k$ of the feature observation in the sliding windows is defined as
\begin{equation}
  {r_k = \frac{1}{m}\sum_{j\neq i}{ \left\| \mathbf{u}_k^{c_i}
    -\pi\left(\mathbf{T}_b^c\mathbf{T}_w^{b_i}\mathbf{T}_{b_j}^w \mathbf{T}_c^b \mathbf{P}_k^{c_j} \right) \right\|},}
\end{equation}
where $\mathbf{u}_k^{c_i}$ is the observation of $k$th feature in the $i$th frame; 
$\mathbf{P}_k^{c_j}$ is the 3D location of $k$th feature in the $j$th frame; 
$\mathbf{T}_c^b$ and $\mathbf{T}_{b_j}^w$ are the transforms from camera frame to body frame and from $j$th body frame to world frame, respecvtively; 
$\pi$ represents the camera projection model.
When the $r_k$ is over a preset threshold, the $k$th feature is considered as a dynamic feature.

As shown in Fig.~\ref{fig:mcc}, the moving consistency check (MCC) module can find out unstable features. However, some stable features are misidentified (top left image), and features on standing people are not recognized (bottom right image). 
A low threshold holds a high recall rate of unstable features.
Further, a misidentified unstable feature with more observations will be recycled if its reprojection error is lower than the threshold.

\begin{table*}[t]
  \caption{Results of RMSE of ATE [$m$], T.RPE [$m/s$], and R.RPE [${}^\circ/s$] ON TUM RGB-D $fr3\_walking$ datasets. }
  \label{table:tum_rgbd}
  \begin{center}
    \begin{tabular}{c|ccc|ccc|ccc|ccc}
      \hline \multirow{2}{*}{ Sequence } 
      & \multicolumn{3}{|c|}{ ORB-SLAM2\cite{mur-artalORBSLAM2OpenSourceSLAM2017} } 
      & \multicolumn{3}{|c|}{ DS-SLAM\cite{yuDSSLAMSemanticVisual2018} } 
      & \multicolumn{3}{|c|}{ Ji $et\ al.$\cite{jiRealtimeSemanticRGBD2021} } 
      & \multicolumn{3}{|c}{ Dynamic-VINS } \\
      & ATE  & T.RPE& R.RPE& ATE  & T.RPE& R.RPE& ATE  & T.RPE& R.RPE& ATE & T.RPE & R.RPE \\
      \hline $fr3\_walking\_xyz$  
      & 0.7521    & 0.4124& 7.7432
      &0.0247&0.0333&0.8266
      &$\mathbf{0.0194}$&$\mathbf{0.0234}$&$\mathbf{0.6368}$
      &0.0486&0.0578 & 1.6932\\
      
      $fr3\_walking\_static$ 
      & 0.3900    & 0.2162 & 3.8958
      &0.0081&0.0102&$\mathbf{0.2690}$
      &0.0111&0.0117&0.2872
      &$\mathbf{0.0077}$&$\mathbf{0.0095}$&0.4581 \\
      
      $fr3\_walking\_rpy$  
      & 0.8705   & 0.4249 & 8.0802
      &0.4442 &0.1503&3.0042
      &$\mathbf{0.0371}$&$\mathbf{0.0471}$&$\mathbf{1.0587}$
      &0.0629&0.0595&5.0839\\
      
      $fr3\_walking\_half$  
      & 0.4863 & 0.3550 & 7.3744
      &0.0303&$\mathbf{0.0297}$&$\mathbf{0.8142}$
      &$\mathbf{0.0290}$&0.0423&0.9650
      &0.0608&0.0665&5.2116 \\
      \hline
    \end{tabular}
  \end{center}
\end{table*}

\begin{table*}[t]
  \caption{{Ablation Experiment Results of RMSE of ATE [$m$], T.RPE [$m/s$], and R.RPE [${}^\circ/s$] ON TUM RGB-D $fr3\_walking$ datasets.}}
  \label{table:tum_ablation}
  \begin{center}
    \begin{tabular}{c|ccc|ccc|ccc|ccc}
      \hline \multirow{2}{*}{ Sequence } 
      & \multicolumn{3}{|c}{ W/O CIRCULAR MASK} 
      & \multicolumn{3}{|c|}{ W/O OBJECT DETECTION}
      & \multicolumn{3}{|c}{ W/O SEG-LIKE MASK} 
      & \multicolumn{3}{|c}{ W/O MCC} \\
      & ATE  & T.RPE& R.RPE
      & ATE  & T.RPE& R.RPE
      & ATE  & T.RPE& R.RPE
      & ATE  & T.RPE& R.RPE \\
      \hline $fr3\_walking\_xyz$ 
      &0.9795&0.6156&6.2692 
      &0.0592&0.0575&1.7181
      &0.0523&0.0608&1.7474
      &0.0676&0.0604&1.8020\\
      
      $fr3\_walking\_static$ 
      &0.4111&0.4052&9.8985 
      &0.3458&0.3136&9.2520  
      &0.0305&0.0194&0.5463
      &0.0454&0.0229&0.5676\\
      
      $fr3\_walking\_rpy$  
      &0.4111&0.4052&9.8985
      &0.2138&0.1191&5.4847 
      &0.1174&0.0729&5.5470
      &0.1236&0.0996&5.4196\\
      
      $fr3\_walking\_half$ 
      &1.1218&0.6779&11.521 
      &0.0988&0.0651&5.1839
      &0.0754&0.0672&5.1952
      &0.1748&0.1169&5.8525 \\
      \hline
    \end{tabular}
  \end{center}
\end{table*}

\section{EXPERIMENTAL RESULTS}\label{experiments}

Quantitative experiments\footnote{The experimental video is available at \url{https://youtu.be/y0U1IVtFBwY}} are performed to evaluate the proposed system's accuracy, robustness, and efficiency.
Public SLAM evaluation datasets, OpenLORIS-Scene \cite{shiAreWeReady2020} and TUM RGB-D \cite{sturmBenchmarkEvaluationRGBD2012}, provide sensor data and ground truth to evaluate SLAM system in complex dynamic environments.
Since our system is built on VINS-Mono\cite{vinsmono} and VINS-RGBD\cite{vinsrgbd}, they are used as the baselines to demonstrate our improvement.
VINS-Mono\cite{vinsmono} provides robust and accurate visual-inertial odometry by fusing IMU preintegration and feature observations.
VINS-RGBD\cite{vinsrgbd} integrates RGB-D camera based on VINS-Mono for better performance.
Furthermore, DS-SLAM\cite{yuDSSLAMSemanticVisual2018} and Ji $et\ al.$\cite{jiRealtimeSemanticRGBD2021}, state-of-the-art semantic algorithms based on ORB-SLAM2\cite{mur-artalORBSLAM2OpenSourceSLAM2017}, are also included for comparison.

The accuracy is evaluated by Root-Mean-Square-Error (RMSE) of Absolute Trajectory Error (ATE), Translational Relative Pose Error
(T.RPE), and Rotational Relative Pose Error (R.RPE).
Correct Rate (CR) \cite{shiAreWeReady2020} measuring the correct rate over the whole period of data is used to evaluate the robustness.
{The RMSE of an algorithm is calculated only for its successful tracking outputs.}
Therefore, the longer an algorithm tracks successfully, the more error is likely to accumulate. 
It implies that evaluating algorithms purely by ATE could be misleading. 
On the other hand, considering only CR could also be misleading. 

In order to demonstrate the efficiency of the proposed system, all experiments of Dynamic-VINS are performed on the embedded edge computing devices, HUAWEI Atlas200 DK and NVIDIA Jetson AGX Xavier. 
And the compared algorithms' results are included from their original papers.
Atlas200 DK has an 8-core A55 Arm CPU (1.6GHz), 8 GB of RAM, and a 2-core HUAWEI DaVinci NPU.
Jetson AGX Xavier has an 8-core ARMv8.2 64-bit CPU (2.25GHz), 16 GB of RAM, and a 512-core Nvidia Volta GPU.
And the results tested on both devices are named Dynamic-VINS-Atlas and Dynamic-VINS-Jetson, respectively.
Yet, to the best of our knowledge, the proposed method is the best-performance real-time RGB-D inertial odometry for dynamic environments on resource-restricted embedded platforms.

\subsection{OpenLORIS-Scene Dataset}
OpenLORIS-Scene\cite{genevaOpenVINSResearchPlatform2020} is a real-world indoor dataset with a large variety of challenging scenarios like dynamic scenes, featureless frames, and dim illumination.
The results on the OpenLORIS-Scene dataset are shown in Fig.~\ref{fig:openloris},
including the results of VINS-Mono, ORB-SLAM2, and DS-SLAM from \cite{genevaOpenVINSResearchPlatform2020} as baselines.

The OpenLORIS dataset includes five scenes and 22 sequences in total.
The proposed Dynamic-VINS shows the best robustness among the tested algorithms.
In $office$ scenes that are primarily static environments, all the algorithms can track successfully and achieve a decent accuracy.
It is challenging for the pure visual SLAM systems to track stable features in $home$ and $corridor$ scenes that contain a large area of textureless walls and dim lighting.
Thanks to the IMU sensor, the VINS systems show robustness superiority when the camera is unreliable.
The scenarios of $home$ and $cafe$ contain a number of sitting people with a bit of motion, and $market$ exists lots of moving pedestrians and objects with unpredictable motion.
And the $market$ scenes cover the largest area and contain highly dynamic objects, as shown in Fig.~\ref{fig:dynamic_feature_recognition}.
Although DS-SLAM is able to filter out some dynamic features, its performance is still unsatisfactory.
VINS-RGBD has a similar performance with Dynamic-VINS in relative static scenes, while VINS-RGBD's accuracy drops in highly dynamic $market$ scenes.
The proposed Dynamic-VINS can effectively deal with complex dynamic environments and improve robustness and accuracy.

\subsection{TUM RGB-D Dataset}

The TUM RGB-D dataset \cite{sturmBenchmarkEvaluationRGBD2012} offers several sequences containing dynamic objects in indoor environments.
The highly dynamic $fr3\_walking$ sequences are chosen for evaluation where two people walk around a desk and change chairs' positions while the camera moves in different motions.
As the VINS system does not support VO mode and the TUM RGB-D dataset does not provide IMU measurements, a VO mode is implemented by simply disabling modules relevant to IMU in Dynamic-VINS for experiments.
The results are shown in Table~\ref{table:tum_rgbd}.
The compared methods' results are included from their original published papers.
The algorithms based on  ORB-SLAM2 and semantic segmentation perform better.
Although Dynamic-VINS is not designed for pure visual odometry, it still shows competitive performance and has a significant improvement over ORB-SLAM2.

{
To validate the effectiveness of each module in Dynamic-VINS, ablation experiments are conducted as shown in Table ~\ref{table:tum_ablation}.
The system without applying circular masks (W/O CIRCULAR MASK) from the Sec.~\ref{sec:feature_tracking} and Sec.~\ref{sec:feature_detect} fails to extract evenly distributed stable features, which seriously degrades the accuracy performance.
Without the object detection (W/O OBJECT DETECTION), dynamic features introduce wrong constraints to impair the system's accuracy.
Dynamic-VINS-W/O-SEG-LIKE-MASK shows the results that mask all features in the bounding boxes.
The background features help the system maintain as many stable features as possible to provide more visual constraints.
The moving consistency check plays an important role when object detection fails, as shown in the column W/O-MCC.
}

\begin{table*}[t]
  \caption{Average Computation time [$ms$] of each module and thread on OpenLORIS {$market$} scenes. }
  \label{table:runtime}
  \begin{center}
    \begin{tabular}{c|c|cc|c|cc|c|c}
      \hline \multirow{2}{*}{ Platforms } &\multirow{2}{*}{ Mehods } 
      & Feature & Feature &Tracking&Dynamic Feature&State&Optimization&Object\\
      & &Tracking & Detection&Thread* &Recognition Modules${}^{\dag}$ &Optimization&Thread* &Detection*\\
      \hline 
      
      \multirow{3}{*}{ \shortstack{HUAWEI\\Atlas200 DK }}
      &VINS-Mono\cite{vinsmono}
      &{18.6226}&{58.2712}
      &{57.6301}
      &- &{76.5047}
      &{85.0247}
      &-\\
      \cline{2-2} 
      
      &VINS-RGBD\cite{vinsrgbd}
      &{20.6066}&{58.9413}
      &{81.4598}
      &-&{75.2211}
      &{83.3476}
      &-\\
      \cline{2-2} 
      
      &Dynamic-VINS
      &{15.5350}&{1.7645}
      &{19.8980}
      &{1.3424}&{74.9509}
      &{82.4916}
      &{17.5850}\\
      \hline

      \multirow{3}{*}{ \shortstack{NVIDIA Jetson\\AGX Xavier }}
      &VINS-Mono\cite{vinsmono}
      &{4.4990} &{14.3691}
      &{10.9123}
      &-
      &{49.5326}
      &{52.4842}
      &-\\
      \cline{2-2} 
      
      &VINS-RGBD\cite{vinsrgbd}
      &{4.1099}&{15.4521}
      &{11.9251}
      &-
      &{49.0472}
      &{52.3388}&-\\
      \cline{2-2}

      &Dynamic-VINS
      &{3.3649}&{0.9396}
      &{5.5416}
      &{0.4707}
      &{43.0424}
      &{47.5377}&{21.9211}\\
      \hline
      
      \multicolumn{9}{l}{{* Tracking Thread, Optimization Thread and Object Detection correspond to the three different threads shown in Fig.~\ref{fig:system_pipeline}, respectively.}}\\
      \multicolumn{9}{l}{$\dag$ Dynamic Feature Recognition Modules sum up the Dynamic Feature Recognition, Missed Detection Compensation, and Moving Consistency}\\
      \multicolumn{9}{l}{\ \ \ Check modules.}
    \end{tabular}
  \end{center}
\end{table*}

\subsection{Runtime Analysis}

This part compares VINS-Mono, VINS-RGBD, and Dynamic-VINS for runtime analysis.
These methods are expected to track and detect 130 feature points, and the frames in Dynamic-VINS are divided into 7x8 grids.
The object detection runs on the NPU/GPU parallel to the CPU.
The average computation times of each module and thread are calculated on OpenLORIS {$market$} scenes; the results run on both embedded platforms are shown in Table~\ref{table:runtime}.
It should be noted that the average computation time is only to be updated when the module is used. 
Specifically, in VINS architecture, the feature detection is executed at a consistent frequency with the state optimization thread, which means the frequency of feature detection is lower than that of Feature Tracking Thread.

On edge computing devices with AI accelerator modules, the single-stage object detection method is computed by an NPU or GPU without costing the CPU resources and can output inference results in real-time.
With the same parameters, Dynamic-VINS shows significant improvement in feature detection efficiency in both embedded platforms and is the one able to achieve instant feature tracking and detection in HUAWEI Atlas200 DK.
The dynamic feature recognition modules (Dynamic Feature Recognition, Missed Detection Compensation, Moving Consistency Check) to recognize dynamic features only take a tiny part of the consuming time.
For real-time application, the system is able to output a faster frame-to-frame pose and a higher-frequency imu-propagated pose rather than waiting for the complete optimization result.

\subsection{Real-World Experiments}

\begin{figure}[htpb]
  \centering
  \includegraphics[width = 0.48\textwidth]{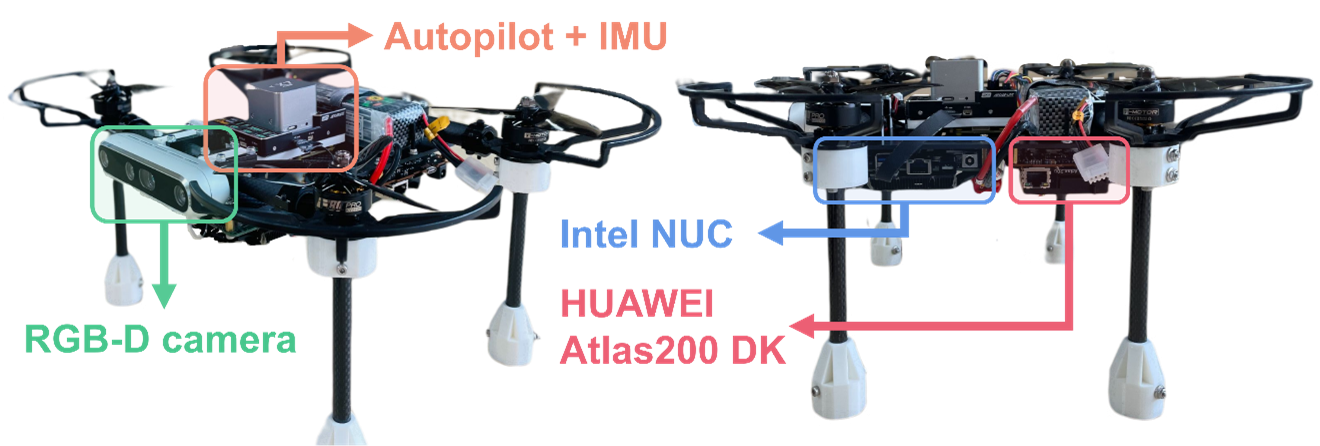}
  \caption{
    A compact aerial robot equipped with an RGB-D camera, an autopilot with IMUs, an onboard computer, and an embedded edge computing device. The whole size is about 255x165mm.
  }
  \label{fig:uav}
\end{figure}

A compact aerial robot is shown in Fig.~\ref{fig:uav}.
An RGB-D camera (Intel Realsense D455) provides 30Hz color and aligned depth images. 
An autopilot (CUAV X7pro) with an onboard IMU (ADIS16470, 200Hz) is used to provide IMU measurements.
The aerial robot is equipped with an onboard computer (Intel NUC, i7-5557U CPU) and an embedded edge computing device (HUAWEI Atlas200 DK).
These two computation resource providers play different roles in the aerial robot.
The onboard computer charges for peripheral management and other core functions requiring more CPU resources, such as planning and mapping.
The edge computing device as auxiliary equipment offers instant state feedback and object detection results to the onboard computer.

\begin{figure}[htbp]
\centering
\begin{minipage}{0.25\textwidth}
  \centering
  \includegraphics[width=\textwidth]{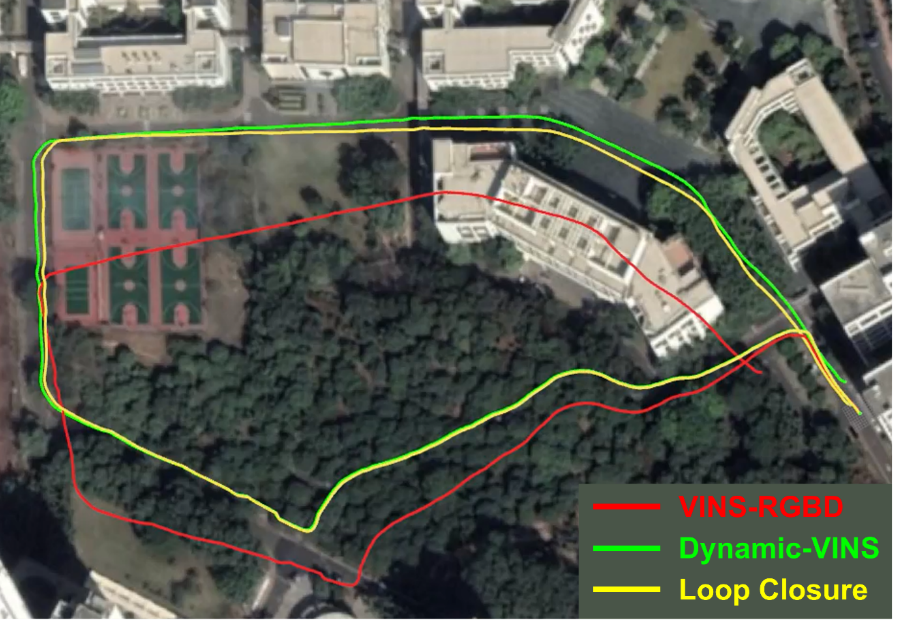}
  \small{(a) HITSZ campus}
\end{minipage}
\begin{minipage}{0.22\textwidth}
  \centering
  \includegraphics[width=\textwidth]{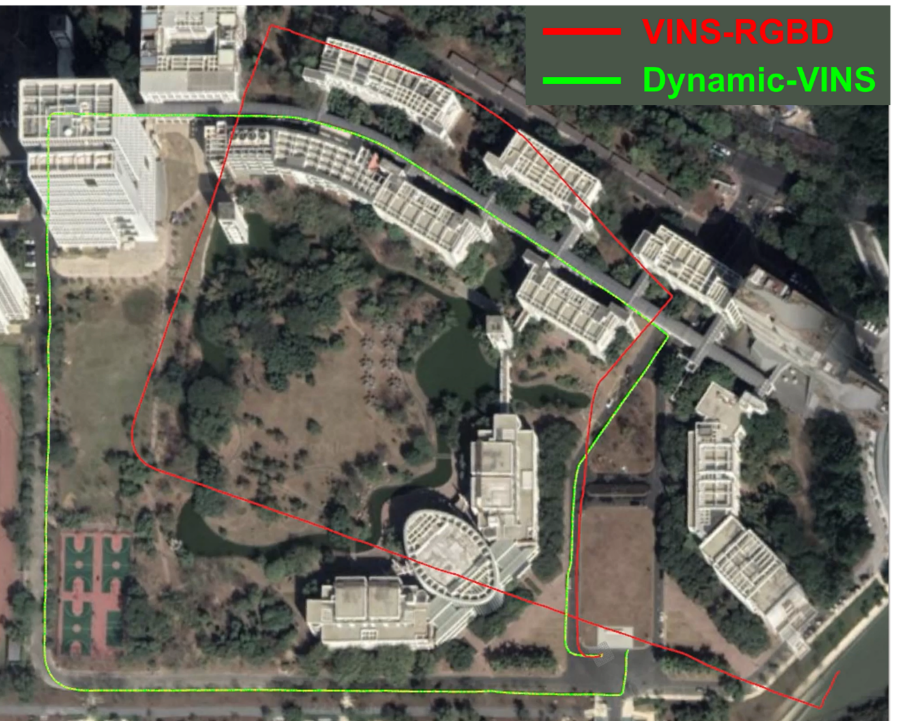}
  \small{{(b) THUSZ campus}}
\end{minipage}
\caption{
  The estimated trajectories in the outdoor environment aligned with the Google map. The green line is the estimated trajectory from Dynamic-VINS, {the red line is from VINS-RGBD,} and the yellow line represents the loop closure that happened at the end of the dataset.
}
\label{fig:google_earth}
\end{figure}

\begin{figure}[htpb]
\centering
\includegraphics[width = 0.48\textwidth]{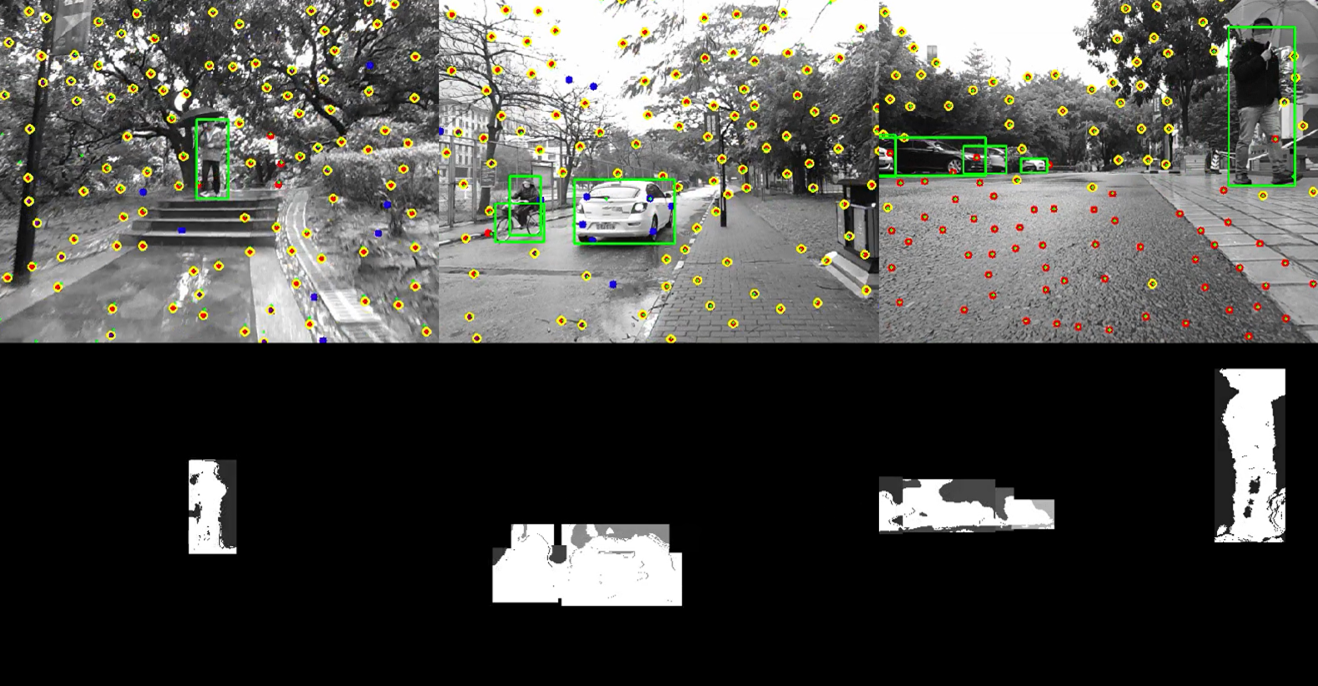}
\caption{
  {Results of dynamic feature recognition in outdoor environments. The dynamic feature recognition modules are still able to segment dynamic objects but with a larger mask region.}
}
\label{fig:outdoor_dynamic_feature_recognition}
\end{figure}

Large-scale outdoor datasets with moving people and vehicles on the HITSZ {and THUSZ }campus are recorded by the handheld aerial robot above for safety.
The total path lengths are approximately $800m$ {and $1220m$, respectively.}
The dataset has a similar scene at the beginning and the end for loop closure, {while loop closure fails in the THUSZ campus dataset}. 
VINS-RGBD and Dynamic-VINS run the dataset on NVIDIA Jetson AGX Xavier. The estimated trajectories and loop closure trajectory aligned with the Google map are shown in Fig.~\ref{fig:google_earth}. 
In outdoor environments, the depth camera is limited in range and affected by the sunlight.
The dynamic feature recognition modules can still segment dynamic objects but with a larger mask region, as shown in Fig.~\ref{fig:outdoor_dynamic_feature_recognition}.
Compared with loop closure results, Dynamic-VINS could provide a robust and stable pose estimation with little drift.

\section{CONCLUSIONS}\label{conclusion}

This paper presents a real-time RGB-D inertial odometry for resource-restricted robots in dynamic environments. 
Cost-efficient feature tracking and detection methods are proposed to cut down the computing burden.
A lightweight object-detection-based method is introduced to deal with dynamic features in real-time.
Validation experiments show the proposed system's competitive accuracy, robustness, and efficiency in dynamic environments.
Furthermore, Dynamic-VINS is able to run on resource-restricted platforms to output an instant pose estimation.
In the future, the proposed approaches are expected to be validated on the existing popular SLAM frameworks. The missed detection compensation module is expected to develop into a moving object tracking module, and semantic information will be further introduced for high-level guidance on mobile robots or mobile devices in complex dynamic environments.


\addtolength{\textheight}{-12cm}

\bibliographystyle{IEEEtran} 
\balance
\bibliography{mybib}




\end{document}